%% file: paper.tex
\documentclass[11pt,a4paper]{article}
\usepackage[hyperref]{acl2018}
\usepackage{times}
\usepackage{latexsym}

\usepackage{url}

\aclfinalcopy 
\def\aclpaperid{1160} 

\usepackage{url}            
\usepackage{booktabs}       
\usepackage{amsfonts}       
\usepackage{nicefrac}       
\usepackage{microtype}      
\usepackage{graphicx}
\usepackage{amsmath}
\usepackage{enumitem}
\usepackage{paralist}
\usepackage{amsmath,amsfonts,amsthm}
\usepackage{color}
\usepackage{array}
\usepackage{epsfig}
\usepackage{caption}
\usepackage{array}
\usepackage{ragged2e}
\usepackage{pbox}
\usepackage{enumitem}
\usepackage{lipsum}
\usepackage[ruled,vlined,boxed,linesnumbered]{algorithm2e}

\usepackage{float}
\restylefloat{table}
\newcolumntype{P}[1]{>{\RaggedRight\hspace{0pt}}p{#1}}
\newcolumntype{L}[1]{>{\raggedright\let\newline\\\arraybackslash\hspace{0pt}}m{#1}}
\newcolumntype{C}[1]{>{\centering\let\newline\\\arraybackslash\hspace{0pt}}m{#1}}
\newcolumntype{R}[1]{>{\raggedleft\let\newline\\\arraybackslash\hspace{0pt}}m{#1}}

\newcommand{\secref}[1]{Section \ref{#1}}
\newcommand{\figref}[1]{Figure \ref{#1}}

\newcommand{\tabref}[1]{Table \ref{#1}}

\newcommand{\cut}[1]{}
\newcommand{\lnear}{\textsc{LocatedNear}}

\begin{document}
	
	\title{Automatic Extraction of Commonsense LocatedNear Knowledge}
	\author{Frank F. Xu\thanks{~~Both authors contributed equally.} ~~~~~~Bill Yuchen Lin\footnotemark[1]~~~~~~Kenny Q. Zhu\\
		Department of Computer Science and Engineering\\
		Shanghai Jiao Tong University \\ Shanghai, China\\ \{frankxu, yuchenlin\}@sjtu.edu.cn, kzhu@cs.sjtu.edu.cn}
	\maketitle
\begin{abstract}
	\lnear~relation is a kind of commonsense knowledge
	describing two physical objects that are typically found near each
	other in real life. 
	In this paper, we study how to automatically extract such relationship through
	a sentence-level relation classifier and aggregating the scores of entity pairs from a large corpus.
	Also, we release two benchmark datasets for evaluation and future research.
\end{abstract}
\input{intro}

\input{framework}

\input{dataset}

\input{eval}

\input{conclude}

\section*{Acknowledgment}
Kenny Q. Zhu is the contact author and was supported by NSFC grants 91646205 and 61373031. Thanks to the annotators for manual labeling, and the anonymous reviewers for valuable comments.

\bibliography{paper}
\bibliographystyle{acl_natbib}
\end{document}

%% file: intro.tex
\section{Introduction}
Artificial intelligence systems can benefit from incorporating commonsense knowledge as background, 
such as \textit{ice is cold} (\textsc{HasProperty}), 
\textit{chewing is a sub-event of eating} (\textsc{HasSubevent}), 
\textit{chair and table are typically found near each other} (\lnear), etc. 
These kinds of commonsense facts have been used in many downstream tasks, such as textual entailment~\cite{dagan2009recognizing,bowman2015large} and visual recognition tasks~\cite{zhu2014reasoning}.
The commonsense knowledge is often represented as relation triples in commonsense knowledge bases, 
such as \textit{ConceptNet}~\cite{speer2012representing}, one of the largest commonsense knowledge graphs available today.
However, most commonsense knowledge bases are manually curated or crowd-sourced by community efforts and thus do not scale well.

\begin{figure}[t]
	\centering
	\includegraphics[width=0.99\columnwidth]{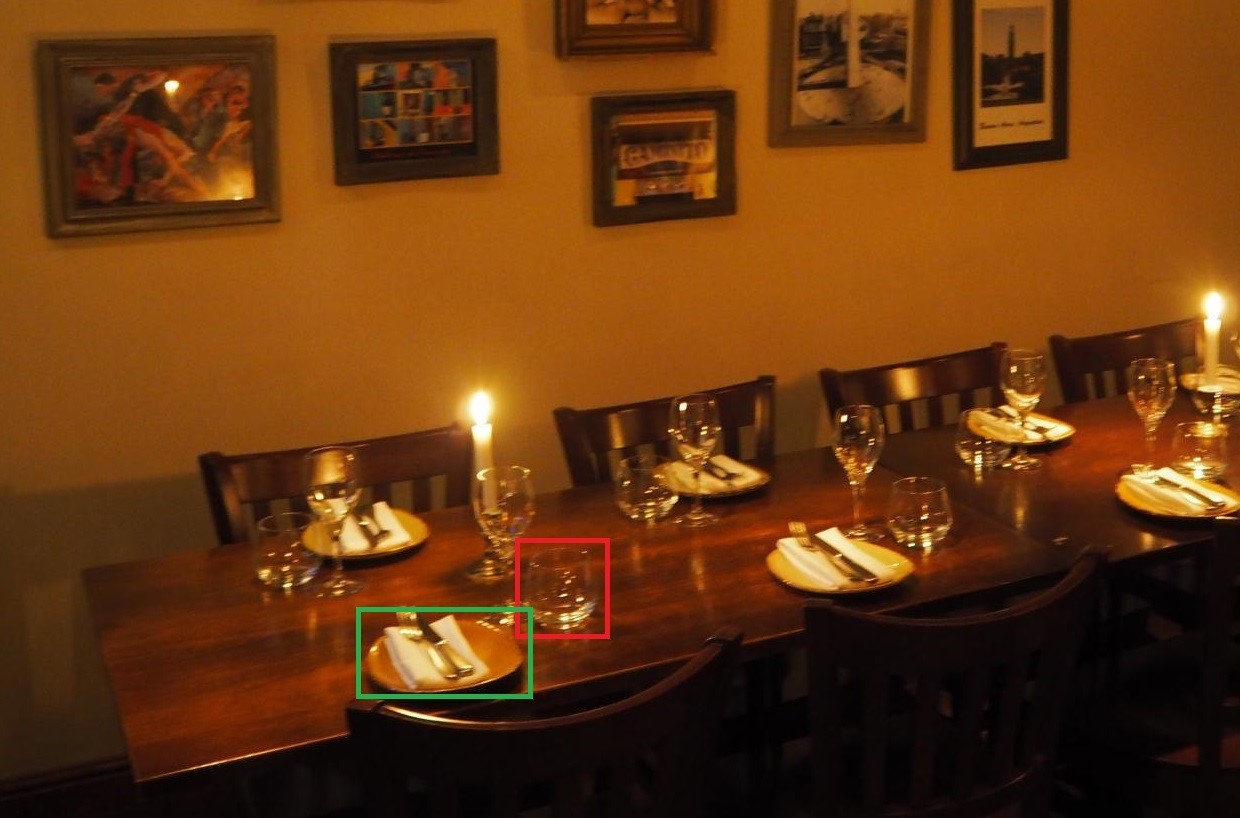}
	\caption{\lnear~ facts assist the detection of vague objects: if a set of knife, fork and plate is on the table, one may believe there is a glass beside based on the commonsense, 
	even though these objects are hardly visible due to low light.}
	\label{fig:dim}
\end{figure}

This paper aims to automatically extract the commonsense \lnear\ relation between
physical objects from textual corpora. \lnear\ is defined as the relationship between two objects typically found near each other in real life.
We focus on \lnear\ relation for these reasons:  
\begin{compactenum}
	\item \lnear\ facts provide helpful prior knowledge to object detection tasks in complex image scenes~\cite{yatskar2016stating}. See~\figref{fig:dim} for an example.
	\item This commonsense knowledge can benefit  reasoning related to spatial facts and physical scenes in reading comprehension,
	question answering, etc.~\cite{li2016commonsense}
	\item Existing knowledge bases have very few facts for	this relation (\textit{ConceptNet 5.5} has only 49 triples of \lnear\ relation).
\end{compactenum}
We propose two novel tasks in extracting \lnear\ relation from textual corpora.
One is a sentence-level relation classification problem which 
judges whether or not
a sentence describes two objects (mentioned in the sentence) being
physically close by.
The other task is to produce a ranked list of \lnear\
facts with the given classified results of 
large number of sentences. 
We believe both two tasks can be used to automatically populate and complete
existing commonsense knowledge bases.

Additionally, we create two benchmark datasets for evaluating \lnear~relation extraction systems on the two tasks: one is 5,000 sentences 
each describing a scene of two physical objects and with a label indicating if the two objects are co-located in the scene; 
the other consists of 500 pairs of objects with human-annotated scores indicating confidences that a certain pair of objects are commonly located near in real life.\footnote{\url{https://github.com/adapt-sjtu/commonsense-locatednear}}

We propose several methods to solve the tasks including feature-based models and LSTM-based neural architectures. 
The proposed neural architecture compares favorably with the current state-of-the-art method for general-purpose relation classification problem.
From our relatively smaller proposed datasets, we extract in total 2,067 new \lnear~triples that are not in \textit{ConceptNet}.

%% file: framework.tex
%
\section{Sentence-level \lnear\ Relation Classification}
\label{sec:classify} 
\textbf{Problem Statement} 
Given a sentence $s$ mentioning a pair of physical objects 
\textless$e_i,e_j$\textgreater, we call \textless$s,e_i,e_j$\textgreater
~an {\em instance}. 
For each instance, the problem is to determine whether $e_i$ and $e_j$ are located near each other in the physical scene described in the sentence $s$.
For example, suppose $e_i$ is ``dog", $e_j$ is ``cat'', and $s$ = ``\textit{The King puts his dog and cat on the table.}''.
As it is true that the two objects are located near in this sentence, a successful classification model is expected to label this instance as \textit{True}.
However, if $s_2$ = ``\textit{My dog is older than her cat.}'', then the label of the instance \textless$s_2,e_i,e_j$\textgreater ~is \textit{False}, because $s_2$ just talks about a comparison in age.
In the following subsections, we present two different kinds of baseline methods for this binary classification task: feature-based methods and  LSTM-based neural architectures.

\subsection{Feature-based Methods}
\label{sec:feature}
Our first baseline method is an SVM classifier based on following features commonly used in many relation extraction models~\cite{xu2015classifying}: 
\begin{compactenum}
	\item  \textit{Bag of Words (BW)}:
	the set of words that ever appeared in the sentence. 
	\item \textit{Bag of Path Words (BPW)}:
	the set of words that appeared on
	the shortest dependency path between objects $e_i$ and $e_j$ in the 
	dependency tree of the sentence $s$, plus the words in the two subtrees 
	rooted at $e_i$ and $e_j$ in the tree.
	\item \textit{Bag of Adverbs and Prepositions (BAP)}:
	the existence of adverbs and prepositions in the sentence as binary features. 
	\item \textit{Global Features (GF)}:
	the length of the sentence, the number of nouns, verbs, adverbs, adjectives, determiners, prepositions and punctuations in the whole sentence. 
	\item \textit{Shortest Dependency Path features (SDP)}:
	the same features as with GF but in dependency parse trees of the sentence and the shortest path 
	between $e_i$ and $e_j$, respectively. 
	\item \textit{Semantic Similarity features (SS)}:
	the cosine similarities between the pre-trained \textit{GloVe} word embeddings~\cite{pennington2014glove} of the two object words.
\end{compactenum}
We evaluate \textit{linear} and \textit{RBF} kernels with different parameter settings, and find the \textit{RBF} kernel with $\{C=100, \gamma=10^{-3}\}$ performs the best overall.

\begin{figure*}[t]
	\centering
	\epsfig{file=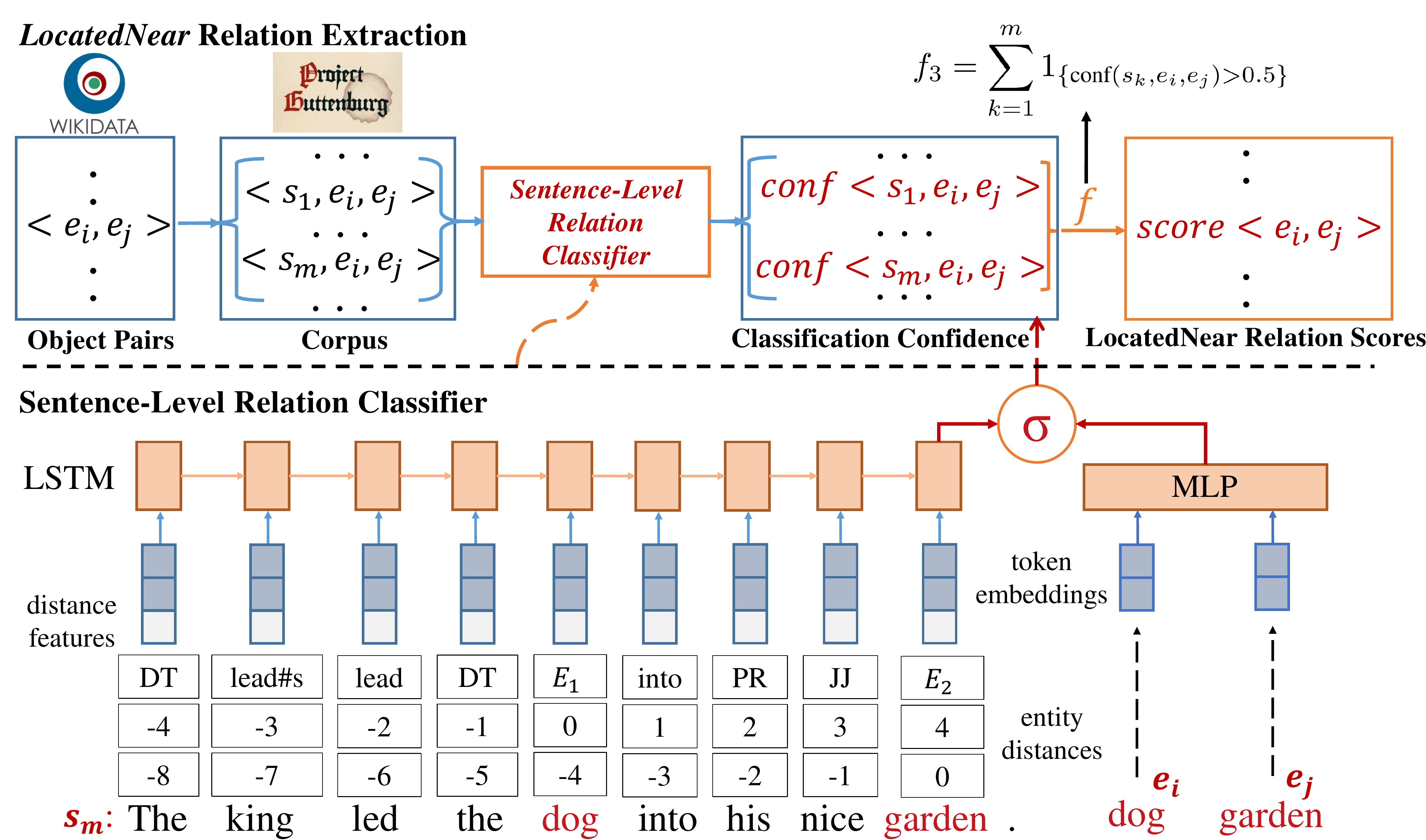, width=1.9\columnwidth}
	\caption{Framework with a LSTM-based classifier}
	\label{fig:LSTM}
\end{figure*}

\subsection{LSTM-based Neural Architectures}
We observe that the existence of \lnear~relation in 
an instance \textless $s$,$e_1$,$e_2$\textgreater~ depends 
on two major information sources: one is from the semantic and syntactical features of sentence $s$ 
and the other is from the object pair \textless$e_1$,$e_2$\textgreater.
By this intuition, we design our LSTM-based model with two parts, shown in lower part of~\figref{fig:LSTM}.
The left part is for encoding the syntactical and semantic information of the sentence $s$, while the right part is encoding the semantic similarity between the pre-trained word embeddings of $e_1$ and $e_2$.

Solely relying on the original word sequence of a sentence $s$ has two problems:
(i) the irrelevant words in the sentence can introduce noise into the model; 
(ii) the large vocabulary of original sentences induce too many parameters, which may cause over-fitting. For example, given two sentences 
	``\textit{The king led the dog into his nice garden.}'' and 
	``\textit{A criminal led the dog into a poor garden.}''. 
	The object pair is \textless \textit{dog, garden}\textgreater~in 
	both sentences.
	The two words ``\textit{lead}'' and ``\textit{into}'' are essential
	for determining whether the object pair is located near, but they 
are not attached with due importance. 
Also, the semantic differences between irrelevant words, such as ``king'' and ``criminal'', ``beautiful'' and ``poor'', are not useful to the co-location
relation between the ``dog'' and ``garden'', and 
thus tend to act as noise.

\begin{table}[t]
	\centering
	\begin{tabular}{l|l}
		\hline
		\textbf{Level}	&  \textbf{Examples}\\ 		\hline
		Objects	& $\textnormal{E}_1$, $\textnormal{E}_2$ \\ 		\hline
		Lemma & open, lead, into, ...\\ \hline 
		Dependency Role	& open\#s, open\#o, into\#o, ... \\ 		\hline 
		POS Tag	& DT, PR, CC, JJ, ... \\ 	\hline 
	\end{tabular}
	\caption{Examples of four types of tokens during sentence normalization. (\#s stands for subjects and \#o for objects)}
	\label{tab:norm}
\end{table}

To address the above issues, we propose a normalized sentence representation method merging the three most important and relevant kinds of information about each instance: lemmatized forms, POS (Part-of-Speech) tags and dependency roles. 
We first replace the two nouns in the object pair as ``$\textnormal{E}_1$'' and ``$\textnormal{E}_2$'', and keep the lemmatized form of the original words for all the \textit{verbs, adverbs and prepositions}, which are highly relevant to describing physical scenes.
Then, we replace the \textit{subjects and direct objects} of the \textit{verbs and prepositions} (\texttt{nsubj, dobj} for verbs and \texttt{case} for prepositions in dependency parse trees) with special tokens indicating their dependency roles. 
For the remaining words, we simply use their POS tags to replace the originals. 
The four kinds of tokens are illustrated in ~\tabref{tab:norm}.
\figref{fig:LSTM} shows a real example of our normalized sentence representation, where the object pair of interest is \textless \textit{dog, garden}\textgreater. 

Apart from the normalized tokens of the original sequence, to capture more structural information, we also encode the distances from each token to $\textnormal E_1$ and $\textnormal E_2$ respectively.
{Such \textit{{position embeddings}} (position/distance features) are proposed by~\cite{zeng2014relation} with the intuition that information needed to determine the relation between two target nouns normally comes from the words which are close to the target nouns.} 


Then, we leverage LSTM to encode the whole sequence of the tokens of normalized representation plus position embedding. 
In the meantime, two pretrained \textit{GloVe} word embeddings~\cite{pennington2014glove} of the original two physical object words are fed into a hidden dense layer. 

Finally, we concatenate both outputs and then use \texttt{sigmoid} activation function to obtain the final prediction.
We choose to use the popular binary cross-entropy as our loss function, and
RMSProp as the optimizer. We apply a dropout rate~\cite{zaremba2014recurrent} of 0.5 in the LSTM and embedding layer to prevent overfitting.

\section{\lnear\  Relation Extraction}
\label{sec:mine}
The upper part of \figref{fig:LSTM} shows the overall workflow of our automatic framework to mine LocatedNear relations from raw text.
We first construct a vocabulary of physical objects and generate all candidate instances. 
For each sentence in the corpus, if a pair of physical objects $e_i$ and $e_j$ appear as nouns in a sentence $s$, then we apply our sentence-level relation classifier on this instance. 
The relation classifier yields a probabilistic score $s$
indicating the confidence of the instance in the existence of~\lnear~relation.
Finally, all scores of the instances from the corpus are grouped by the object pairs and aggregated, where each object
pair is associated with a final score. 
These mined physical pairs with scores can easily be integrated into existing commonsense knowledge base.

More specifically, for each object pair \textless$e_i,e_j$\textgreater, 
we find all the $m$ sentences in our corpus mentioning both objects.
We classify the $m$ instances with the sentence-level relation classifier and obtain confidence scores for each instance, then
feed them into a heuristic scoring function $f$ to obtain the final aggregated score for the given object pair. 
We propose the following 5 choices of $f$ considering accumulation and threshold:
\begin{align}
	f_0&=m\\
	f_1&=\sum_{k=1}^{m}\textnormal{conf}(s_k,e_i,e_j)\\
	f_2&=\frac{1}{m}\sum_{k=1}^{m}\textnormal{conf}(s_k,e_i,e_j)\\
	f_3&=\sum_{k=1}^{m}1_{\{\textnormal{conf}(s_k,e_i,e_j)>0.5\}} \\
	f_4&=\frac{1}{m}\sum_{k=1}^{m}1_{ \{ \textnormal{conf}(s_k,e_i,e_j)>0.5 \}}
\end{align}

%% file: dataset.tex
\section{Datasets}
\label{sec:data}
Our proposed vocabulary of single-word physical objects is constructed by the intersection of all ConceptNet concepts and all entities that belong to ``physical object'' class in \textit{Wikidata}~\cite{VK:wikidata14}. 
We manually filter out some 
words that have the meaning of an abstract concept, which 
results in 1,169 physical objects in total.

Afterwards, we utilize a cleaned subset of the Project Gutenberg corpus~\cite{lahiri:2014:SRW}, which contains 3,036 English books written by 142 authors.
An assumption here is that sentences in fictions are more likely to describe real life scenes. 
We sample and investigate the density of \lnear~ relations in Gutenberg with other 
widely used corpora, namely \textit{Wikipedia}, 
used by~\citeauthor{mintz2009distant}~(2009) and \textit{New York Times} corpus~\cite{riedel2010modeling}. 
In the English \textit{Wikipedia} dump, out of all sentences which mentions at least two
physical objects, 32.4\% turn out to be positive. 
In the \textit{New York Times} corpus,
the percentage of positive sentences is only 25.1\%. 
In contrast, that percentage in the Gutenberg corpus is 55.1\%, much higher 
than the other two corpora, making it a good choice for \lnear~ 
relation extraction.

From this corpus, we identify 15,193 pairs that co-occur in more than 10 sentences.
Among these pairs, we randomly select 500 object pairs and 
10 sentences with respect to each pair for annotators to label their commonsense~\lnear. 
Each instance is labeled by at least three annotators who are college students
and proficient with English. 
The final truth labels are decided by majority voting. 
The Cohen's Kappa among the three annotators is 0.711 which suggests substantial agreement~\cite{Landis1977TheMO}. 
This dataset has almost double the size of those most
popular relations in the SemEval task~\cite{sem}, and the sentences in our
data set tend to be longer.
{We randomly choose 4,000 instances as the training set and 1,000 as the test set for evaluating the sentence-level relation classification task.}
For the second task, we further ask the annotators to label whether each pair of objects are likely to locate near each other in the real world. 
Majority votes determine the final truth labels.
The inter-annotator agreement here is {0.703} (substantial agreement).  

%% file: eval.tex
\section {Evaluation}
\label{sec:eval}
\begin{table*}[th]
	\centering
	\small
	\begin{tabular}{|c|c|c|c|c|c|c|c|}
			\hline
			& Random       & Majority    & SVM  & SVM(-BW)    & SVM(-BPW) & SVM(-BAP)   & SVM(-GF)    \\ \hline
			Acc.  & 0.500        & 0.551       & 0.584                                & 0.577          & 0.556        & 0.563          & \textbf{0.605} \\ \hline
			P & 0.551        & 0.551       & 0.606                                & 0.579          & 0.567        & 0.573          & \textbf{0.616} \\ \hline
			R    & 0.500        & 1.000       & 0.702                                & 0.675          & 0.681        & \textbf{0.811} & 0.751          \\ \hline
			F1        & 0.524        & 0.710       & 0.650                                & 0.623          & 0.619        & 0.672          & \textbf{0.677} \\ \hline \hline
			& SVM(-SDP) & SVM(-SS) & DRNN & LSTM+Word    & LSTM+POS   & LSTM+Norm    &                \\ \hline
			Acc.  & 0.579        & 0.584       & 0.635                                & 0.637          & 0.641        & \textbf{0.653} &                \\ \hline
			P & 0.597        & 0.605       & \textbf{0.658}                       & 0.635          & 0.650        & 0.654          &                \\ \hline
			R    & 0.728        & 0.708       & 0.702                                & \textbf{0.800} & 0.751        & 0.784          &                \\ \hline
			F1        & 0.656        & 0.652       & 0.679                                & 0.708          & 0.697        & \textbf{0.713} &                \\ \hline
		\end{tabular}
	\caption{Performance of baselines on co-location classification task with ablation. (Acc.=Accuracy, P=Precision, R=Recall, ``-'' means without certain feature)}
	\label{tab:aprf}
\end{table*}
In this section, we first present our evaluation of our proposed methods and the state-of-the-art general relation classification model on the 
first task.  
Then, we evaluate the quality of the new \lnear~triples we extracted.

\subsection{Sentence-level \lnear\ Relation Classification}
We evaluate the proposed methods against the state-of-the-art general domain relation 
classification model (DRNN)~\cite{Xu2016ImprovedRC}. 
The results are shown in~\tabref{tab:aprf}.
For feature-based SVM, we do feature ablation on each of the 6 feature types. For LSTM-based model, we experiment on variants of input sequence of original sentence:
``LSTM+Word'' uses the original words as the input tokens;
``LSTM+POS'' uses only POS tags as the input tokens; 
``LSTM+Norm'' uses the tokens of sequence after sentence normalization. 
Besides, we add two naive baselines: ``Random'' baseline method
classifies the instances into two classes with equal probability. 
``Majority'' baseline method considers all the instances to be positive.

From the results, we find that the SVM model without the Global Features performs best, which indicates that bag-of-word features benefit more in shortest dependency paths than on the whole sentence.
Also, we notice that DRNN performs best (0.658) on precision but not significantly higher than LSTM+Norm (0.654). 
The experiment shows that LSTM+Word enjoys the highest recall score, while
LSTM+Norm is the best one in terms of the overall performance. 
One reason is that the normalization representation reduces the vocabulary
of input sequences, while also preserving important syntactical 
and semantic information.  
Another reason is that the~\lnear\ relation are described in sentences decorated
with prepositions/adverbs.
These words are usually descendants of the object word in the dependency tree, 
outside of the shortest dependency paths. 
Thus, DRNN cannot capture the information from the words belonging to 
the descendants of the two object words in the tree, 
but this information is well captured by LSTM+Norm. 

\subsection{\lnear\ Relation Extraction}
Once we have obtained the probability score for each instance using LSTM+Norm, we can extract \lnear\
relation using the scoring function $f$. 
We compare the performance of 5 different heuristic choices of $f$, by quantitative results. 
We rank 500 commonsense \lnear\ object pairs described in \secref{sec:mine}. \tabref{tab:3m} shows the ranking results using
\textit{Mean Average Precision} (MAP) and \textit{Precision} at $K$ as the metrics. 
Accumulative scores ($f_1$ and $f_3$) generally do better. 
Thus, we choose $f = f_3$ with a MAP score of 0.59 as the scoring function.
\begin{table}[t]
	\centering
	\small
	\begin{tabular}{|cccccc|}
		\hline
		${f}$	& {MAP} & {P@50} & {P@100}  &  {P@200}& {P@300}\\ \hline \hline
		$f_0$ & 0.42 & 0.40 & 0.44 & 0.42 & 0.38 \\ \hline
		$f_1$	& 0.58  & {\bf 0.70} & 0.60& 0.53 & {\bf 0.44}\\\hline
		$f_2$	& 0.48 & 0.56 & 0.52  & 0.49 & 0.42\\\hline
		$f_3$	& {\bf 0.59} & 0.68& {\bf 0.63} & {\bf 0.55} & {\bf 0.44}\\\hline
		$f_4$	& 0.56 & 0.40 & 0.48 & 0.50 & 0.42\\\hline
	\end{tabular}
	\caption{Ranking results of scoring functions.}
	\label{tab:3m}
\end{table}

\begin{table}[th]
	\centering
	\small
	\begin{tabular}{|ccc|}
		\hline
		(door, room)  & (boy, girl)     & (cup, tea)      \\
		(ship, sea)   & (house, garden) & (arm, leg)      \\
		(fire, wood)  & (house, fire)   & (horse, saddle) \\
		(fire, smoke) & (door, hall)    & (door, street)  \\
		(book, table) & (fruit, tree)   & (table, chair)  \\ \hline
	\end{tabular}
	\caption{Top object pairs returned by best performing scoring function $f_3$}
	\label{tbl:toppairs}
\end{table} 

Qualitatively, we show 15 object pairs with some of the highest $f_3$ scores
in \tabref{tbl:toppairs}.
Setting a threshold of 40.0 for $f_3$, which is the minimum non-zero
$f_3$ score for all true object pairs in the \lnear\ object pairs 
data set (500 pairs), we obtain a total of 2,067 \lnear\ relations, with
a precision of 68\% by human inspection.

%% file: conclude.tex
\section{Conclusion}
In this paper, we present a novel study on enriching \lnear\ relationship from textual corpora. 
Based on our two newly-collected benchmark datasets, we propose several methods to solve the sentence-level relation classification problem. 
We show that existing methods do not work as well on this task and discovered that LSTM-based model does not have significant edge over simpler feature-based model. 
Whereas, our multi-level sentence normalization turns out to be useful.

Future directions include: 1) better leveraging distant supervision to reduce human efforts, 2) incorporating knowledge graph embedding techniques, 3) applying the \lnear\ knowledge into downstream applications in computer vision and natural language processing.